\documentclass[runningheads]{llncs}
\usepackage{graphicx}
\usepackage{amsmath,amssymb} 
\usepackage{enumitem}
\usepackage{multirow}
\usepackage[table, dvipsnames]{xcolor}
\usepackage{color}
\usepackage{cite}
\usepackage{setspace}
\usepackage{url}

\begin{document}

\title{Artistic Object Recognition by\\Unsupervised Style Adaptation}
\titlerunning{Artistic Object Recognition by Unsupervised Style Adaptation} 

\author{Christopher Thomas \and Adriana Kovashka}

\authorrunning{Thomas and Kovashka} 

\institute{University of Pittsburgh, Pittsburgh PA 15260, USA \\
\email{\{chris,kovashka\}@cs.pitt.edu}}

\maketitle

%%%%%%%%% ABSTRACT
%\vspace{-1.5em}
\begin{abstract}
Computer vision systems currently lack the ability to reliably recognize artistically rendered objects, especially when such data is limited. In this paper, we propose a method for recognizing objects in artistic modalities (such as paintings, cartoons, or sketches), without requiring any labeled data from those modalities. Our method explicitly accounts for stylistic domain shifts between and within domains. To do so, we introduce a complementary training modality constructed to be similar in artistic style to the target domain, and enforce that the network learns features that are invariant between the two training modalities. We show how such artificial labeled source domains can be generated automatically through the use of style transfer techniques, using diverse target images to represent the style in the target domain. Unlike existing methods which require a large amount of unlabeled target data, our method can work with as few as ten unlabeled images. We evaluate it on a number of cross-domain object and scene classification tasks and on a new dataset we release. Our experiments show that our approach, though conceptually simple, significantly improves the accuracy that existing domain adaptation techniques obtain for artistic object recognition.
\end{abstract}
%%%%%%%%% BODY TEXT
%\vspace{-2.5em}
\section{Introduction}
\label{sec:intro}

\begin{figure}[t]
    \centering
    \includegraphics[width=0.95\linewidth]{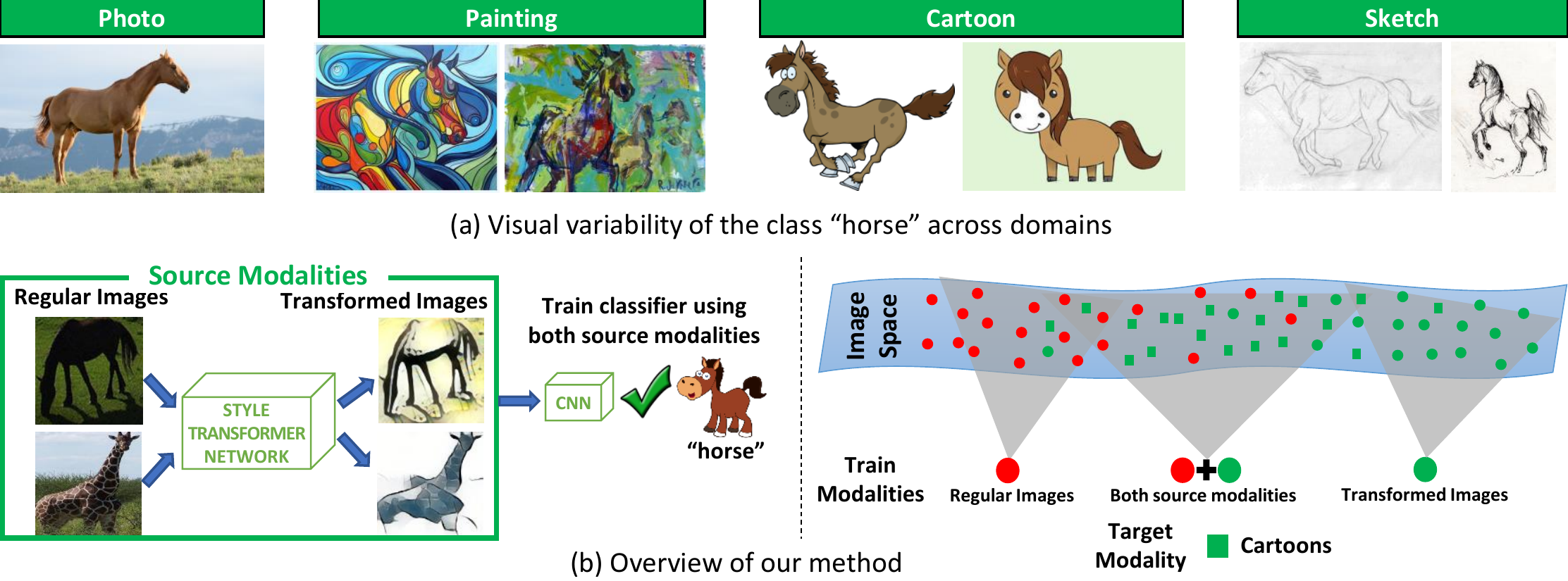}
    %\vspace{-2em}
    \caption{(a) Illustration of the visual variation between instances of the same object  across domains. (b) Overview of our method: We transform our training data so it resembles the style of the target domain. Left: We illustrate how we use style modification to create a synthetic source modality that looks like the target modality, causing the learned features to be more target-appropriate. Right: Training on any single modality causes learned features to be \emph{too} domain specific, so we train on both domains.}
    \label{fig:concept}
    %\vspace{-1.5em}
\end{figure}

Clever design of convolutional neural networks, and the availability of large-scale image datasets for training \cite{lin2014microsoft,russakovsky2015imagenet},
have greatly increased the performance of object recognition systems. 
However, models trained on one dataset often do not perform well on another \cite{oquab2014learning}. For example, training a model on photographs and applying it to an artistic modality such as cartoons is unlikely to yield acceptable results given the large differences in object appearance across domains \cite{li2017deeper}, as illustrated in Fig.~\ref{fig:concept} (a).

One possible solution to this problem is to obtain labeled training data from 
the domain that the model will be applied on (i.e. the \textit{target} modality). 
Unfortunately, obtaining sufficient data for training today's deep 
convolutional networks on each modality of interest may be prohibitively 
expensive or impossible. In our scenario of recognition in artistic modalities, this problem is especially pronounced: a given artist usually does not produce limitless samples for a particular object class, yet each artist may have a unique style. This problem has led researchers to explore \emph{domain adaptation} techniques: a model is trained on a domain where data is plentiful (the \textit{source} domain) 
while ensuring that what the model learned can be applied to a disjoint \emph{target} domain. 
Usually these approaches rely on a small amount of labeled data in the target domain \cite{tzeng2015simultaneous} or on seeing some \emph{unlabeled} target data and learning a feature space where the distributions on the source and target domains are similar \cite{gong2012geodesic,ganin2015unsupervised,bousmalis2016domain,gupta2016cross,ghifary2016deep,long2016unsupervised, Tzeng_2017_CVPR, Bousmalis_2017_CVPR}.
 
In contrast to the more general domain adaptation methods above, we focus on a \emph{specific type of domain difference}. We consider domains which exhibit differences due to artistic style \cite{gatys2016image,johnson2016perceptual,Huang_2017_ICCV}, such as paintings, cartoons and sketches.
Note that artistic domains often contain limited data---e.g. the PACS dataset of \cite{li2017deeper} contains on average less than 2,500 images per domain.

We propose two variants of our method, both of which learn style-invariant representations for recognition, one using as few as ten unlabeled target images.  
Our methods construct a new, \emph{synthetic source modality} from a single photorealistic modality. The synthetic modality bears the \emph{style} of the target modality, and is obtained for free (without human annotations) through the use of style transfer techniques \cite{johnson2016perceptual,Huang_2017_ICCV}.
We then train a network using both source modalities, ensuring that the features the model learns produce similar activations in the two source domains.
Unlike existing methods which also create synthetic training data for domain adaptation \cite{Bousmalis_2017_CVPR,Tzeng_2017_CVPR}, our method is easier to train, and requires orders of magnitude less target data (e.g. ten vs ten thousand). This translates to more accurate performance for domains where even unlabeled target data is sparse.
We illustrate our method in Fig.~\ref{fig:concept} (b). 

Our method of generating synthetic data in the same artistic style as the target domain is applicable to any setting where the primary difference between the domains is the style of the content, e.g.\ cartoons, paintings, sketches, and line drawings. It is not however applicable to problems where the source and target domain differences extend beyond artistic style, e.g. RGB-D data.

We evaluate our approach on multiple datasets (both new and from prior work) and show it outperforms the state-of-the-art domain adaptation methods. 

The main contributions of our work are as follows:
\begin{itemize}[itemsep=3pt,topsep=3pt,parsep=0pt,partopsep=0pt]
    \item We propose a framework for constructing synthetic source modalities useful for learning style-invariant classifiers through style transfer.
    \item We develop a novel style selection component, which enables accurate adaptation with just ten target images.
     \item We release a new large dataset of photographs and artistic images across four domains: natural photographs, cartoons, paintings, and sketches. The dataset is available at \url{http://www.cs.pitt.edu/~chris/artistic_objects/}.
	\item The two versions of our approach outperform recent domain adaptation techniques on the challenging task of object recognition in artistic modalities, which exhibits larger domain shift than traditional adaptation scenarios.
    \item We conduct rich evaluation against many baselines in different settings (also see our supplementary file).
\end{itemize}

%\vspace{-1em}
\section{Related Work}
\label{sec:related}
%\vspace{-0.1em}
\textbf{Recognition on artistic modalities.}
There is limited work in performing object recognition on artistic modalities.
\cite{ginosar2014detecting} show that as paintings become more abstract, the performance of person detection degrades. 
\cite{wu2014learning,cai2015beyond} benchmark existing methods on artistic objects, but only propose improvements over older, non-convolutional techniques, which are greatly outperformed by neural networks.
\cite{wilber2017bam} provide a large dataset of artistic domains with per-image object annotations, but we found the labels were too coarse (due to multiple objects in the same image and no bounding box annotations) and the human annotations too sparse, to reliably perform recognition.
\cite{li2017deeper} publish a dataset of objects in artistic modalities, and propose a domain \emph{generalization} approach. We use this dataset but instead focus on domain \emph{adaptation}, where we target a specific unlabeled modality, rather than attempting to do well on an unseen modality.
Further, \cite{li2017deeper} require the availability of multiple original (non-synthetic) source modalities, while we only require one real modality. 

%\vspace{-0.3cm}
\textbf{Cross-domain retrieval.}
There has been interest in retrieving samples across domain boundaries, e.g. retrieving photographs that match a \emph{sketch}.
\cite{sangkloy2016sketchy} find a joint embedding for sketches and photographs that ensures a paired sketch and photo are more similar in the learned space than a photo and sketch that are not paired. 
We utilize the dataset of \cite{sangkloy2016sketchy} but perform recognition and not retrieval, and do not assume availability of labeled data from the target domain. 
\cite{Castrejon_2016_CVPR} retrieve scene types across e.g. sketches and clipart. 
They propose a supervised approach (which we modify to an unsupervised setting for comparison) that encourages source and target neural activations to fit the same distribution.

%\vspace{-0.3cm}
\textbf{Domain adaptation.}
Domain adaptation techniques can be broadly divided into two categories: 1) semi-supervised  \cite{tzeng2015simultaneous,long2015learning} in which a limited amount of labeled target data is available, and 2) unsupervised techniques \cite{gong2012geodesic,ganin2015unsupervised,ghifary2016deep,long2016unsupervised,Bousmalis_2017_CVPR} (such as this paper) where only unlabeled target data is available.
Older domain adaptation techniques either modify a classifier or ensure that the feature space in which we attempt to classify a novel modality are similar to the features for the source modality \cite{gong2012geodesic}.
Recently, researchers have tackled the same problem in the context of training convolutional neural networks.
\cite{tzeng2015simultaneous,ganin2015unsupervised} show how to learn domain-invariant features with CNNs by confusing domains, i.e. ensuring that a classifier cannot guess to which domain an image belongs based on its activations. 
Other approaches  \cite{ghifary2016deep,bousmalis2016domain} train an autoencoder 
which captures features specific to the target domain,  
then train a classifier on the source and apply it to the target. 
While these approaches attempt to bring the \emph{source and target} domains closer, we encourage feature invariance between our \emph{two source} domains, while explicitly bringing the source and \emph{target} closer via style transfer. 
We compare against a number of recent, state-of-the-art unsupervised domain adaptation approaches and find that our approach 
consistently outperforms them.

Several works use \emph{multiple source domains} for domain adaptation 
but these assume multiple real human-labeled sources are available \cite{gupta2016cross,li2017deeper} or can be discovered in the given data \cite{gong2013reshaping}. 
Recent works \cite{Bousmalis_2017_CVPR,Tzeng_2017_CVPR, HoffmanTPZISED18} have tackled unsupervised domain adaptation using generative adversarial networks (GANs) \cite{goodfellow2014generative} to generate synthetic target data. 
The networks are trained to produce novel data such that a discriminator network cannot tell it apart from the real target dataset. \cite{HoffmanTPZISED18} extends the idea further by training a GAN which adapts source data at both the image and feature level. 
Our method has several important advantages over GAN-based approaches. We exploit the fact that our target modality's domain gap with respect to the source can be bridged by controlling for artistic style. Since we know we want to model style differences, we can explicitly extract and distill them, rather than requiring a GAN to learn these in an unsupervised way. This allows our method to use \textit{orders of magnitude less target data}. 
The ability to leverage limited target data is essential in applications where target data is extremely limited, such as artistic domains \cite{li2017deeper}. Second, training generative adversarial networks is challenging \cite{salimans2016improved} as the training is unstable and dataset-dependent. Our image translation networks are \textit{easier to train} and require no tuning. 
Our approach outperforms \cite{Bousmalis_2017_CVPR} in experiments.

Other domain adaptation works have also attempted to model style. \cite{zhang2013writer} propose a framework for handwriting recognition from different writers which projects each writer's writing into a ``style-free'' space and uses it to train a classifier. Another classic work \cite{tenenbaum1997separating} separates style and content using bilinear models and can generate a character in a certain writing style.
Unlike our proposed methods, neither of these works use CNNs or perform object recognition.

%\vspace{-0.3cm}
\textbf{Domain generalization.}
Domain generalization methods \cite{muandet2013domain,xu2014exploiting,li2017domain,niu2015visual,li2017deeper} attempt to leverage knowledge from one or more source domains during training, and apply it on completely unseen target domains. 
\cite{li2017deeper} proposes a method of domain generalization on a dataset of paintings, cartoons, and sketches. In order to generalize to cartoons, for example, they assume data is available from all other modalities. In contrast, we perform adaptation towards a known, but unlabeled target modality, without requiring multiple source modalities.

%\vspace{-0.3cm}
\textbf{Style transfer.}
Style transfer methods \cite{gatys2016image,johnson2016perceptual,Huang_2017_ICCV,chen2017stylebank, huang2017real, Chen_2017_ICCV} modify an image or video to have the style of some single target image. For example, we might modify a portrait to have the same artistic style as ``Starry Night'' by Van Gogh. \cite{gatys2016image} use a pixel-by-pixel loss, whereas \cite{johnson2016perceptual} use a more holistic perceptual loss to ensure that both content and style are preserved while an input image is transformed. 
While \cite{gatys2016image} use a backpropagation technique to iteratively transform an image to another style, \cite{johnson2016perceptual} train a CNN to directly perform the style transfer in a feed-forward network.
More recently, \cite{Huang_2017_ICCV} propose an encoder-decoder framework which allows  mimicking the style of a target image without retraining a network. 
Earlier work \cite{thomas2016seeing} models style at the object part and scene level.
We use \cite{johnson2016perceptual,Huang_2017_ICCV} as the style transfer component of our framework.
Importantly, the transformed data retains the classification labels from the photograph source data.

%\vspace{-1em}
\section{Approach}
\label{sec:approach}
%\vspace{-0.25em}
Our method explicitly controls for the stylistic gap between the source and target domains.
While our method is based on existing  style transfer approaches, its novelty is two-fold. 
First, we explicitly encourage variation in the styles towards which the source modality is modified, which we call \emph{style selection}. In Sec.~\ref{sec:results} and our supplementary file, we show that this novel style selection is crucial for attaining good performance in practice.
Second, while other methods proposed to use synthetic data for adaptation, we apply our approach to a new problem: object recognition in \emph{artistic} domains. 
This problem is characterized by larger domain shift compared to prior adaptation benchmarks, as shown in \cite{li2017deeper}; and by sparse unlabeled target data (see dataset sizes in Sec.~\ref{sec:datasets}).

The main intuition behind our approach is that we know a large part of the domain gap between photographs and our target modalities is their difference in artistic style. The key observation is that most of the images within an artistic modality exhibit one of a few ``representative'' styles. For example, some paintings may be cubist, expressionist, or photorealistic. Sketches may be different due to low-level detail, but look similar in terms of coarse visual appearance. Other unsupervised methods \cite{Bousmalis_2017_CVPR,ghifary2016deep} require a large dataset of target images to learn this shared appearance. 
Because we specifically focus on artistic style variations, 
we can explicitly distill ``representative'' styles from clusters of artistically similar images, and then transfer those styles onto our labeled training data. We train on both original and transformed photos, and add a constraint to our method to encourage the features on both domains to be indistinguishable. Because the only difference between the real and synthetic domains is their style, this constraint, perforce, \textit{factors out artistic style} from our feature representation and bridges the domain gap. 
Below we first describe how we obtain additional source modalities, and then how we learn style-invariant features during training.

\begin{figure}[t]
%\vspace{-0.25cm}
    %\resizebox{1\linewidth}{!}{
    \includegraphics[width=1\linewidth]{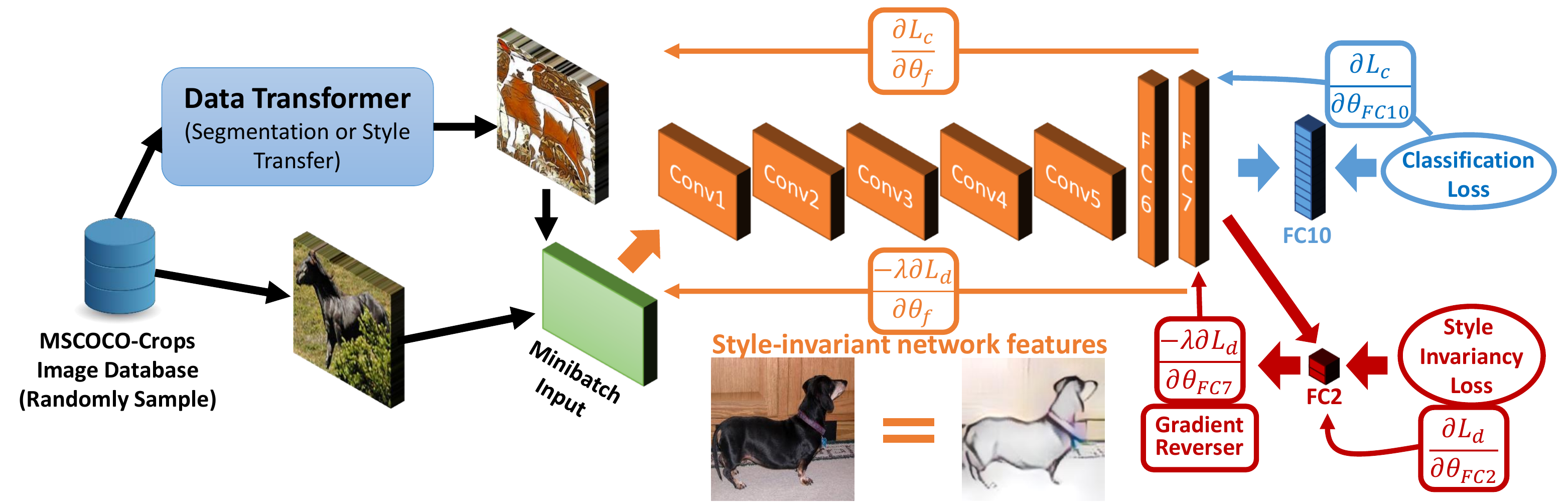}
    %}
    %\vspace{-2.5em}
    \caption{Training with multiple modalities and style-invariance constraint. We train networks on real and synthetic data. We show an example of style transfer transforming photos into labeled synthetic cartoons. The style-invariance loss trains the FC2 layer to predict which modality the image came from. During backpropagation, we reverse its gradient before propagating it to the layers used by both classifiers. This encourages those layers to learn style-invariant features.}
    \label{fig:method}    
    %\vspace{-1.25em}
\end{figure}

Our approach automatically transforms photographs into the representative styles found in the target modality. Because the photos are labeled, the transformed photos retain their original labels.
We consider two techniques for style transfer. While they are comparable in terms of perceptual quality overall, one \cite{johnson2016perceptual} produces much higher-quality sketches (see the supplementary file), but is computationally less efficient, which limits the number of transformations we can create. We enable this method to produce rich, albeit few, transformations using a novel representative \textit{style selection} technique (Sec.~\ref{sec:style_selection}). The second style transfer technique \cite{Huang_2017_ICCV} can be used to create more transformations as it is more efficient, but it produces less stable results. 

%\vspace{-1.2em}
\subsection{Style transfer via Johnson \cite{johnson2016perceptual}}
\label{sec:johnson}
%\vspace{-0.25em}

Johnson~et~al.~\cite{johnson2016perceptual} present an artistic style transfer technique which trains a feed-forward network to perform transfer towards \textit{a single style}. Thus, performing transfer towards additional styles requires training multiple networks (which is computationally expensive).
The style transformation network is trained to minimize two losses: the source loss $\mathcal{S}$, which preserves the content of a given source image, and the target loss $\mathcal{T}$, which preserves the style of a given (unlabeled) target image. Let $\Phi_i(I)$ be the neural activations of layer $i$ of a pretrained VGG-16 \cite{simonyan2014very} network on input image $I$. Let $I_s$ denote a source domain image, $I_t$ a target domain image, and $\widehat{I_s}$ a source domain image transformed by the style transfer network. The shape of $\Phi_i(I)$ is given by $C_i \times H_i \times W_i$, which represents the number of channels, height, and width of the activations of layer $i$ respectively. The source loss $\mathcal{S}$ is thus given by:
%\vspace{-0.4em}
\begin{equation}
\label{eq:source}
\mathcal{S}^{\Phi} = \sum_i \frac{1}{C_i H_i W_i} \left\lVert \Phi_i(\widehat{I_s}) - \Phi_i(I_s) \right\rVert_2^2 
\end{equation}
%\vspace{-0.75em}

\noindent 
In practice, we use only a single VGG layer to compute our source loss, $i=$ \verb|relu3_3|, based on the experimental results of \cite{johnson2016perceptual}. 

The target loss, on the other hand, maintains the ``style'' of the target image, preserving colors and textures. 
Style spatial invariance is achieved by taking the distance between the correlations of different filter responses, by computing the Gram matrix $G^{\Phi_i}(I)$, where the matrix entry $G_{a,b}^{\Phi_i}(I)$ is the inner product between the flattened activations of channel $C_{i_a}$ and $C_{i_b}$ from $\Phi_i(I)$. 
Our target loss is then just the normalized difference between the Gram matrices of the transformed source image and the target image:

%\vspace{-1em}
\begin{equation}
\mathcal{T}^{\Phi} = \sum_i \left\lVert G^{\Phi_i}(\widehat{I_s}) - G^{\Phi_i}(I_t) \right\rVert_{F}^2
\end{equation}
%\vspace{-1em}

\noindent where $F$ represents the Frobenius norm. Following \cite{johnson2016perceptual}'s results, we use $i \in $ \{\texttt{relu1\_2, relu2\_2, relu3\_3, relu4\_3}\} for our target loss. The final loss formulation for our style transformation network is then $\mathcal{L} = \lambda_s \mathcal{S}^{\Phi} + \lambda_t \mathcal{T}^{\Phi}$, where $\lambda_s=1$ and $\lambda_t=5$. 
We omit the total variation regularizer term  for simplicity.

%\vspace{-1em}
\subsubsection{Style selection. } 
\label{sec:style_selection}
In order to capture the artistic style variation \textit{within} our target dataset, we choose a set of representative style images from our dataset to represent clusters with the same artistic style. We run the target data through layers in the style loss network that are used in computing the style reconstruction loss. We then compute the Gram matrix on the features to remove spatial information. Because of the high dimensionality of the features, we perform PCA and select the first thousand components. We cluster the resulting vectors using $k$-means, with $k=10$. From each ``style cluster'' we select the image closest to its centroid as its style representative. We then train one style transfer network for each style representative, as described above.

\begin{figure}[t]
    \centering
     \includegraphics[width=1\linewidth]{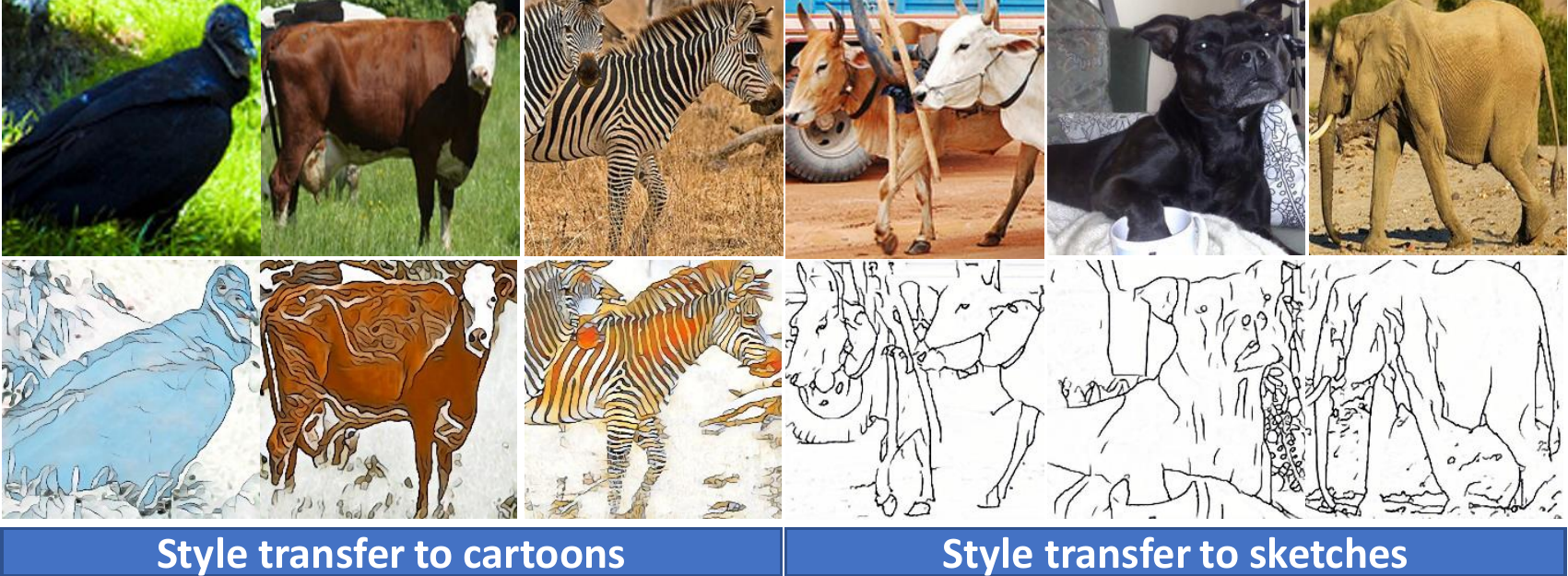}
    \caption{Examples of COCO images transformed towards two modalities: cartoons (left) and sketches (right). Background information is lost, which might be useful in non-photorealistic modalities such as cartoons that may not even have a background.}
    \label{fig:style}
\end{figure}

%\vspace{-1em}
\subsection{Style transfer via Huang \cite{Huang_2017_ICCV}}
\label{sec:huang}
%\vspace{-0.3em}
Because style transfer via the previous approach requires training a network for every style representative, it is too computationally expensive for more than a few representative styles. We thus also explore a second, more recent style transfer technique by Huang and Belongie \cite{Huang_2017_ICCV}. The model follows an hourglass, encoder-decoder architecture. The key takeaway of the approach is that style transfer can be performed on the output of the encoder (i.e. in feature space) by transferring channel-wise mean and variance across spatial locations. Using the same notation as above, style transfer is performed by transforming the encoder's output as follows, where $\widehat{\Phi_i(I_s)}$ represents the transformed source image features, and $\mu$ and $\sigma$ represent the mean and variance across all spatial locations in the given feature map:
%\vspace{-0.25em}
\begin{equation}
\widehat{\Phi_i(I_s)} = \sigma\left( \Phi_i(I_t) \right) \left(  \frac{\Phi_i(I_s) - \mu\left( \Phi_i(I_s) \right)}{\sigma\left( \Phi_i(I_s) \right)}  \right) + \mu \left(  \Phi_i(I_t)  \right) 
\end{equation}
Notice that after training the decoder reconstruction network (the encoder is a pre-trained VGG network, as in Johnson \cite{johnson2016perceptual}), performing style transfer towards any arbitrary image only requires one to compute $\Phi_i(I_s)$ and $\Phi_i(I_t)$, transform the source image features, and provide $\widehat{\Phi_i(I_s)} $ to the decoder to get the style-transferred image. This enables style transfer to be performed towards any arbitrary target image in a feedforward, efficient manner, without any retraining.

%\vspace{-0.75em}
\subsection{Creating the synthetic source modality}
\label{sec:style_transfer}

Formally, let $\mathbf{I}_s = \left\{ I^i_s, y^i_s \right\}_{i=1}^{N_s}$ denote the dataset of labeled source data and $\mathbf{I}_t = \left\{ I^i_t \right\}_{i=1}^{N_t}$ the dataset of unlabeled target data, where $N_s$ and $N_t$ represent the number of images in each dataset. Our style transfer network $\Psi(I^i_s, \theta_t^j) \rightarrow \widehat{I^i_s}$, transforms each source domain image to a synthetic image $\widehat{I^i_s}$ which appears in the style of the single target image $j$ captured by the parameters $\theta_t^j$.  In the case of Johnson's method \cite{johnson2016perceptual}, we train one style-transfer network for each of ten ``style representatives,'' and apply each learned network to modify all source images. In the case of Huang's method \cite{Huang_2017_ICCV}, we randomly select ten target images \textit{for each source image} $I^i_s$ as our target styles. The transformed dataset is thus more stylistically diverse for our adaptation approach via Huang's method. For each target image sampled for either method, we obtain a transformed dataset in the style of that target, $\widehat{\mathbf{I}^{I^j_t}_s} = \left\{ \Psi(\mathbf{I}_s, \theta_t^j), \mathbf{y}_s \right\}$, where $\mathbf{y}_s$ denote object labels from the source data. We emphasize the targets are only used for their style, \textit{not their content} (i.e. not their object labels). 
Examples of style modification for cartoons and sketches are shown in Fig.~\ref{fig:style}. 

%\vspace{-1em}
\subsection{Training with two modalities}
\label{sec:multiple_modality_training}

We train convolutional neural networks which are exposed to two modalities of data.
We modify the training procedure such that each minibatch passed to the network during training contains
data from both real and synthetic domains. 
In order to explicitly control for style differences between the source and target domains in our learning process, we impose a style-invariance constraint on the network. The constraint causes the network to learn a feature representation such that it is difficult to predict which of the two source modalities an image came from, given the network's features.
In other words, the network learns to \emph{confuse or perform transfer between the source modalities}. 

The key observation is that because we have deliberately constructed the synthetic modality such that it differs from the real modality \textit{only in its artistic style}, in order to learn modality-indistinguishable features across both modalities, the network is forced to \textit{factor-out} style from its representation and thus learn style-invariant features for recognition.
A similar criterion is used in prior work \cite{ganin2015unsupervised, tzeng2015simultaneous} but to confuse the \emph{source and target modalities}.
This is an important distinction: because the source and target modalities may differ in several ways, 
learning modality-invariant features between them may cause the network to disable filters which ultimately are useful for classification on the target domain. 
In our case of source-to-source invariance, because our source domains contain the same images except in different styles, the network is only forced to disable filters which depend on artistic style.
We illustrate our method in Fig.~\ref{fig:method}.

For each modality, we train a CNN using our source and transformed data to perform the following two classification tasks: 
%\vspace{-0.65em}
\begin{equation}
T\left( \left\{ \mathbf{I}_s, \widehat{ \mathbf{I}_s } \right\};  \theta_Y; \theta_F; \theta_D \right) \rightarrow \left\{ \mathbf{y}, \mathbf{d} \right\}
\end{equation}
%\vspace{-1.5em}

\noindent where $\widehat{ \mathbf{I}_s }$ denotes the collection of transformed data output by our style transfer networks. 
The first classification task (predicting $\mathbf{y}$) predicts the object class of the image using the parameters $\theta_Y$. The second task is to predict from which modality a given image came from (denoted by $\mathbf{d}$) using the parameters $\theta_D$. Both classifiers make their predictions using the intermediate network features $\theta_F$. 
Thus, we seek the following parameters of the network: $\theta_F$ which \textit{maximizes} the loss of the modality classifier while simultaneously minimizing the loss of the object classifier, $\theta_Y$ which \textit{minimizes} the object classification loss, and $\theta_D$ which \emph{minimizes} the modality classifier's loss. Our optimization process seeks the saddle point of the following adversarial functional:

%\vspace{-0.6em}
\begin{equation}
\label{eq:full_method}
    \min_{\theta_F,\theta_Y} \max_{\theta_D} \; \alpha \mathcal{L}_d \left( \left\{ \mathbf{I}_s, \widehat{\mathbf{I}_s} \right\}, \mathbf{d} \right) + \beta \mathcal{L}_y \left( \left\{ \mathbf{I}_s, \widehat{\mathbf{I}_s} \right\}, \mathbf{y} \right)
\end{equation}
%\vspace{-0.8em}

\noindent where $\alpha$ and $\beta$ are weighting parameters. In practice, we set $\alpha$ to be $1/10$ the value of $\beta$. We use a multinomial logistic loss to train both classifiers. $\mathcal{L}_d$ represents our domain classification loss,
\noindent while $\mathcal{L}_y$ is the object classifier loss which trains our network to recognize the object categories of interest.

\noindent \textbf{Implementation Details:} We experiment exclusively with the AlexNet architecture in this paper, but we include results with ResNet in our supplementary. We train all CNNs using the Caffe \cite{jia2014caffe} framework. We use a batch size of 256 for AlexNet 
and a learning rate of $\alpha=$1e-4; higher learning rates caused training to diverge.
We used SGD with momentum $\mu = 0.9$ and learning rate decay of $\gamma=0.1$ every 25K iterations. We trained all networks for a maximum of 150K iterations, but found the loss typically converged much sooner.

%\vspace{-1.25em}
\section{Experimental Validation}

%\vspace{-0.6em}
\subsection{Datasets}
\label{sec:datasets}
%\vspace{-0.4em}

Most of our experiments focus on object recognition, using the \textbf{PACS} dataset of \cite{li2017deeper} and the \textbf{Sketchy Database} of \cite{sangkloy2016sketchy} as our target domains.
We complement these with our own dataset, described below.
To test our method with a broader set of categories, we also test on \textbf{Castrejon} et al.'s \cite{Castrejon_2016_CVPR} large dataset containing 205 scene categories in different modalities.
Our method also uses a photorealistic modality as the original source modality; this is either the photo domain from the respective dataset, or COCO \cite{lin2014microsoft} (for our new dataset). 

\textbf{PACS} \cite{li2017deeper} contains four domains, three artistic (sketches, cartoons, and paintings) and one photorealistic, which we use as the original, non-synthetic source domain. The dataset contains seven object categories (e.g. ``dog'', ``elephant'', ``horse'', ``guitar'') 
and 9,991 images in total. 
The authors demonstrate the much larger shift exhibited in PACS compared to prior adaptation datasets, by showing KL divergence and cross-domain accuracy loss statistics.

\textbf{The Sketchy Database} \cite{sangkloy2016sketchy} contains 75,471 human-drawn sketches and accompanying photographs, from 125 common object categories in ImageNet.

\textbf{Castrejon}  \cite{Castrejon_2016_CVPR} extends the Places dataset \cite{zhou2014learning} containing 205 scene categories by providing four new modalities: sketches, clipart, text descriptions, and spatial text images. We use the two \emph{visual} modalities, sketches and clipart.

\textbf{New dataset: CASPA.}
Since the PACS dataset \cite{li2017deeper} is small, we complemented it by assembling images in the same three modalities as PACS. We use this additional dataset to test our conclusions on more test data. 
We call the dataset CASPA: (\underline{Ca}rtoons, \underline{S}ketches, \underline{Pa}intings).
We assembled a dataset of 5,047 cartoons by querying Google Image Search with each of the ten animal categories from COCO: ``bear'', ``bird'', ``cat'', ``cow'', ``dog'', ``elephant'', ``giraffe'', ``horse'', ``sheep'', and ``zebra''. We chose these categories because sketch data of these animal classes was available.
We also collected a new dataset of how different painters paint  objects. We downloaded 1,391
paintings\footnote{from http://www.arab-painting.com/pic/Oil Painting Styles on Canvas/Animals} which cover eight of our ten categories (except ``giraffe'' and ``zebra''), and annotated these images with \emph{2,834 bounding boxes.} 
Note that another painting dataset exists \cite{cai2015beyond} but it only contains four of the ten classes we consider.
To maintain the same set of domains as those used in PACS, we also include 12,008 sketches from the Sketchy Database \cite{sangkloy2016sketchy}.
PACS uses seven categories from \cite{sangkloy2016sketchy}, while we use twenty categories and collapse them down into coarser categories (i.e. different types of birds became the coarse category ``bird'').
In total, the dataset contains 18,446 images, almost twice more than PACS' 9,991.
The dataset is available at \url{http://www.cs.pitt.edu/~chris/artistic_objects/}.
As our photorealistic source domain, we use COCO \cite{lin2014microsoft}.
Because images in COCO contain multiple objects, we use the provided bounding boxes to produce cropped images containing a single object, and eliminate crops which do not contain the ten animal categories listed above.

%\vspace{-1em}
\subsection{Baselines}
\label{sec:baselines}

We compare against several recent unsupervised domain adaptation methods; these do not require any labeled data on the target domain.

\begin{itemize}[leftmargin=*,itemsep=3pt,topsep=3pt,parsep=0pt,partopsep=0pt]
\item \textsc{Gong} et al. \cite{gong2012geodesic} is a pre-CNN method which shows how statistical differences between features extracted from two domains can be corrected, by learning the geodesic flow between the two domains. We use FC7 features.
\item \textsc{Ganin} and Lempitsky \cite{ganin2015unsupervised} make CNNs  modality-invariant by learning features such that the CNN is unable to distinguish which modality the image came from, via a domain confusion loss. Since no labeled data is available from the target, we only use \cite{ganin2015unsupervised}'s domain confusion loss but not the classification loss.
\item \textsc{Ghifary} et al. \cite{ghifary2016deep} train an encoder-decoder architecture to compute the reconstruction loss with the target domain's data, and use the resulting features to train a classifier on the labeled source data.
\item \textsc{Long} et al. \cite{long2016unsupervised} propose to train separate classifiers for the source and target domain which differ only by a residual function. The source and target classifiers can still preserve domain-specific cues useful for classifying in that domain, but the source classifier is expressed as a residual function of the target classifier, and its residual layers capture the source-only features. 
\item \textsc{Castrejon} et al. \cite{Castrejon_2016_CVPR} learn a representation across scenes, by fixing the higher layers of the network while the lower layers are allowed to learn a domain-dependent representation. However, unlike \cite{Castrejon_2016_CVPR}, we cannot learn modality-specific lower-layer features without target labels.
We train on ImageNet and fine-tune the higher layers on our photo labeled data, but we skip the second part of \cite{Castrejon_2016_CVPR}'s ``Method A'' since we do not have labels on the modality data; we instead use generic pretrained ImageNet lower-level features.
\item \textsc{Bousmalis} et al. \cite{Bousmalis_2017_CVPR} use a generative adversarial network (GAN) to transform labeled source data to appear as if it were target data. The generator adapts source images to fool the discriminator, while the discriminator attempts to distinguish between the generator's output and real target images. A classifier is jointly trained to classify the original and translated source images.
\end{itemize}

We also compare against \textsc{Photo-AlexNet}, a standard AlexNet \cite{krizhevsky2012imagenet}, trained on 1.3M ImageNet samples, fine-tuned on our photorealistic domain (COCO / Places, respectively). See our supplementary file for results using ResNet.

Note that we do not compare to the domain \emph{generalization} approach in \cite{li2017deeper} since it assumes the availability of \emph{multiple} human-labeled source modalities at training time, while we only require one. For \textsc{Bousmalis} et al.\ and \textsc{Ghifary} et al.\ we use data generated by their methods instead of our style-transfered data, then proceed with training a network with two modalities and domain confusion loss. We found this made these methods more competitive.

\begin{table}[ht]
\centering
\small 
\resizebox{1\linewidth}{!}{
\begin{tabular}{|c||c|c|c||c||c|c|c||c||c|}
\hline
& \multicolumn{9}{c|}{\textbf{Target Domain}}\\
\hline
\hline
 & \multicolumn{4}{c|}{\textbf{PACS}} & \multicolumn{4}{c|}{\textbf{CASPA}} & \textbf{Sketchy} \\   
\hline
\textbf{Method} 								& \textbf{Paint} 		& \textbf{Cart} & \textbf{Sketch} 	& \textbf{AVG} 			
		& \textbf{Paint} 	& \textbf{Cart} 	& \textbf{Sketch} 	& \textbf{AVG} 			
        & \textbf{Sketch} \\
\hline
\hline
\textsc{Photo-AlexNet} 							& 0.560 			& 0.276 		& 0.328 			& 0.388 					& 0.663 		& 0.222 			& 0.398 			& 0.428 		& 0.093\\
\hline
\hline
\textsc{Gong} \cite{gong2012geodesic} 			& 0.487 			& 0.310 		& 0.305 			& 0.367 					& 0.309 		& 0.200 			& 0.463 			& 0.324 		& 0.112\\
\hline
\textsc{Ganin} \cite{ganin2015unsupervised} 	&\textbf{0.624} 	& 0.\emph{638} 	& 0.351 			& 0.538 					& 0.652 		& 0.251 			& 0.202 			& 0.368 		& 0.112\\ 
\hline
\textsc{Ghifary} \cite{ghifary2016deep} 		& 0.453 			& 0.561 		& 0.371 			& 0.462 					& 0.631		& 0.388 			& 0.418		&  	0.479	& 0.225\\ 
\hline
\textsc{Long} \cite{long2016unsupervised} 		& 0.614				&\textbf{0.668}	& 0.417				& 0.566 					& 0.628 		& \emph{0.425} 			& 0.497 			& 0.517 		& \emph{0.303}\\
\hline
\textsc{Castrejon} \cite{Castrejon_2016_CVPR} 	& 0.580				& 0.293			& 0.328				& 0.400 					& 0.628 		& 0.231 			& 0.458 			& 0.439 		& 0.085\\
\hline
\textsc{Bousmalis} \cite{Bousmalis_2017_CVPR} 	& 0.609 			& 0.472 		& 0.559 			& 0.547 					& 0.666		& 0.408 			& 0.482 				& 0.519			& 0.284\\
\hline
\hline
\textsc{Ours-Johnson} 				&\textbf{0.624} 	& 0.485 		& \emph{0.653} 		& \emph{0.587} 							& 	\emph{0.677}			& 0.406 		& \textbf{0.625}		& 		\textbf{0.569}		& \textbf{0.326}\\
\hline
\textsc{Ours-Huang}					& 0.619 			& 0.480 		&\textbf{0.689} 	&\textbf{0.596} 						& \textbf{0.698} 		& \textbf{0.464} 	& \textit{0.501} 			& \emph{0.554} 		& 0.234\\
\hline
\hline
\textsc{Upper Bound}  							& 0.863 			& 0.927 		& 0.921 			& 0.904 					& 0.842 		&  0.741					&  	0.917 	& 		0.833		& 0.822\\
\hline
\end{tabular}
}
%\vspace{0.5em}
\caption{Our method outperforms all other domain adaptation techniques on average. The best method is in \textbf{bold} (excluding upper bound), and the second-best in \emph{italics}.} 
\label{tab:main}
%\vspace{-3em}
\end{table}

\subsection{Results}
\label{sec:results}

In Table \ref{tab:main}, we show the results of our comparison to state-of-the-art unsupervised domain adaptation methods. 
We first show the result of \emph{not} performing any adaptation (\textsc{Photo-AlexNet}). We then show the performance of six domain adaptation techniques (from 2012, 2015, 2016, and 2017). 
We next show the two versions of our method, using style transfer techniques from Johnson (Sec.~\ref{sec:johnson}) and Huang (Sec.~\ref{sec:huang}).
Finally, we show the performance of an \textsc{Upper bound} method which ``cheats'' by having access to labeled target data at training time, which none of the baselines nor our methods see.  
We split the data from each modality into 90\%/10\% training/test sets (for the sake of the upper bound which requires target data).
All results show top-1 accuracy.

\textbf{Our results.} 
We see that one of our methods is always the best performer per dataset (average columns for \textbf{PACS} and \textbf{CASPA}, and the single column for \textbf{Sketchy}), and the other is second-best in two of three cases. For cartoons and paintings, \textsc{Ours-Huang} is the stronger of our two methods, or our two methods perform similarly. On the full \textbf{Sketchy} dataset, which \textbf{PACS-Sketch} and \textbf{CASPA-Sketch} are subsets of, \textsc{Ours-Johnson} performs much better (by about 40\%). This is because \cite{Huang_2017_ICCV} and thus \textsc{Ours-Huang} has trouble generating high-quality sketch transfer. 
This tendency holds for the subset chosen in \textbf{CASPA} (as \cite{johnson2016perceptual} is better than \cite{Huang_2017_ICCV} by about 25\%), but on the subset in \textbf{PACS}, \cite{Huang_2017_ICCV} does somewhat better (6\%). We believe this is due to the choice of categories in each sketch subset; we illustrate these results in our supplementary.
Note \textbf{Sketchy} contains many more classes hence the lower performance overall.

While we observe \emph{some} qualitative correlation between visual quality of the style transferred images and utility of the data for training our classifiers, we note that our key concern in this work is utility, not visual quality. We do observe, however, that in some cases loss of detail in synthetic sketches may be causing our classifiers trained on such data to struggle with classification, particularly for \textsc{Ours-Huang}. We explore this problem in detail in our supplementary material. 

\textbf{Baseline results.}
Our style transfer approach outperforms all other domain adaptation methods on all types of data except \textbf{PACS-Painting} (where it ties for best performance with \textsc{Ganin}), and on \textbf{PACS-Cartoon}. 
Our method outperforms the strongest baseline by 23\% for \textbf{PACS-Sketch} (wrt \textsc{Bousmalis}), 
5\% for \textbf{CASPA-Painting} (wrt \textsc{Bousmalis}), 
9\% for \textbf{CASPA-Cartoon} (wrt \textsc{Long}), and 25\% for \textbf{CASPA-Sketch} (wrt \textsc{Long}). On average, \textsc{Ours-Johnson} and \textsc{Ours-Huang} outperform the best baseline by 4\%/5\% respectively on \textbf{PACS} (wrt \textsc{Long}), and 10\%/7\% on \textbf{CASPA} (wrt \textsc{Long}). 
\textsc{Ours-Johnson} outperforms \textsc{Long} by 8\% for \textbf{Sketchy}. To underscore the importance of style adaptation as opposed to na\"ive data transformations, we show using style transferred data outperforms edge maps (which are sketch-like) on sketches in our supp. file.

Interestingly, on \textbf{CASPA-Painting}, only three domain adaptation methods outperform the no-adaptation photo-only model, but our two are the only ones that do so by a significant margin. As shown in \cite{li2017deeper}, paintings are closer to photographs than cartoons and sketches are, so it is not surprising that several domain adaptation methods fail to improve upon the source model's performance. By explicitly accounting for the style of the target modality, we improve the photo model's accuracy by 5\%.

%\vspace{-1em}
\textbf{Data requirements.} 
Overall, the strongest baseline is \textsc{Long}.
The most recent method, \textsc{Bousmalis} \cite{Bousmalis_2017_CVPR}, performs second-best.
The performance of this GAN model is limited  by the amount of training data available. While \cite{Bousmalis_2017_CVPR} uses all unlabeled target samples, that set of target samples is too limited to adequately train a full generative adversarial network. As a point of comparison, in their original work, \cite{Bousmalis_2017_CVPR} use 1000 \textit{labeled} target images just to verify their hyperparameters. This is 100 times the amount of target data we use for domain adaptation in \textsc{Ours-Johnson}. This result underscores one of the key strengths of our approach: its ability to explicitly distill and control for style while requiring little target data.
Notice that for all domain adaptation methods, there is still a large gap between the unsupervised recognition performance, and the supervised upper bound's performance.

%\vspace{-1em}
\textbf{Ablations.} 
Our method is a framework for domain adaptation that explicitly accounts for the artistic style variations between domains. Within this framework, we show how to make the best use of limited target data, via the style selection technique in Sec.~\ref{sec:johnson}. In Sec.~\ref{sec:multiple_modality_training}, we encourage style-invariant features to be learned, via a domain confusion loss over the source domains. We briefly demonstrate the benefit of these two techniques here; see the supplementary file for the full ablation study.

Style selection of ten diverse target style images has a great advantage over randomly choosing ten images for transfer. On \textbf{PACS-Sketches}, style selection achieves a 15\% improvement over randomly choosing the style images (0.653 vs 0.569) and it achieves 14\% improvement on \textbf{PACS-Cartoons} (0.485 vs 0.425). While style selection does require access to the target dataset to \textit{choose} the representative styles from, our technique still significantly improves performance when a single set of ten images are selected at random (i.e. when the target data pool is limited to ten images); see our supp.\ for more details.

Encouraging style invariance via the domain confusion loss also boosts performance, by 28\% (0.384 for style transfer using random targets and domain confusion loss, vs 0.299 without that loss) for \textbf{CASPA-Cartoons}, and 16\% for \textbf{CASPA-Sketches} (0.594 vs 0.513).

% ----------------------------------------

%\vspace{-1em}
\textbf{Results on Castrejon's dataset.} 
In Table \ref{tab:places}, we evaluate our methods against the two strongest baselines from Table \ref{tab:main}, this time on the scene classification task in \cite{Castrejon_2016_CVPR}'s dataset.
Note that unlike \cite{Castrejon_2016_CVPR}, we do not train on target labels, and we only train with 250K Places images.
Performance is low overall on this 205-way task. 
While our methods' performance is closely followed by \textsc{Bousmalis}, our methods perform best. In particular, \textsc{Ours-Johnson} outperforms both \textsc{Bousmalis} and \textsc{Long} on both modalities. This experiment confirms our conclusions from Table \ref{tab:main} that explicitly controlling for style variation is key when the domain shift is caused by artistic style.

\begin{table}[t]
\centering
\small
\begin{tabular}{|c|c|c|}
\hline
\textbf{Method} 							& \textbf{Castrejon-Clipart} & \textbf{Castrejon-Sketches} 	\\ \hline \hline
\textsc{Photo-AlexNet} 						&  				0.0689	&  	0.0213	\\ \hline \hline
\textsc{Long} \cite{long2016unsupervised} 	& 	0.0727	& 		0.0402			\\ \hline 
\textsc{Bousmalis} \cite{Bousmalis_2017_CVPR} 	&  		0.0839			&  		\emph{0.0464}				\\ \hline \hline

\textsc{Ours-Johnson} 						&  		\emph{0.0847}	& \textbf{0.0479} 				\\ \hline
\textsc{Ours-Huang}							&		\textbf{0.0885}					& 0.0456								\\ \hline \hline
\textsc{Upper Bound}						&		0.5937					&		0.3120		\\ \hline 
\end{tabular}
\caption{Scene classification results using \cite{Castrejon_2016_CVPR}'s data.}
\label{tab:places}
%\vspace{-2em}
\end{table}

%\vspace{-1em}
\textbf{Supplementary file.} Please see our supplementary for extensive additional experiments showing the performance of ablations of our method (comparing style selection vs.\ selecting random style images), methods that use a single source modality (including the synthetic modalities generated by our method), domain-agnostic style transfer using edge maps, a comparison of methods using the ResNet architecture, and a detailed exploration of the differences between the Huang \cite{Huang_2017_ICCV} and Johnson \cite{johnson2016perceptual} style transfer methods which help determine which method is appropriate for a given target domain.

%\vspace{-1em}
\section{Conclusion}
\label{sec:conclusion}
%\vspace{-0.5em}

We addressed the problem of training CNNs to recognize objects in unlabeled artistic modalities. 
We found the introduction of style-mimicking training modalities generated for free 
results in a domain-invariant representation. 
We confirm in extensive experiments that our method outperforms a variety of state-of-the-art and standard baselines. Compared to recent works which create synthetic training data, our method requires far less target data (as few as ten unlabeled target images) and is easier to train. We also release a large dataset in photographs and images showing objects across multiple artistic domains.
Currently our method cannot handle large, exaggerated shape differences found in some cartoons or sketches. In future work, we will investigate making our domain adaptation method even more robust by performing more aggressive transformations, so we can recognize objects in even more abstract artistic renderings.

%\vspace{-0.8em}
%\begin{spacing}{0.88}
\noindent \textbf{Acknowledgement:} This material is based upon work supported by the National Science Foundation under NSF CISE Award No.\ 1566270.
\bibliographystyle{splncs04}
\bibliography{references}
%\end{spacing}
\end{document}